\documentclass[10pt,twocolumn,letterpaper]{article}

\usepackage{iccv}
\usepackage{times}
\usepackage{epsfig}
\usepackage{graphicx}
\usepackage{amsmath}
\usepackage{amssymb}
\usepackage{graphicx}
\usepackage{caption}
\usepackage{adjustbox}
\usepackage{multirow}
\usepackage{float}
\usepackage{amsmath}
\usepackage{color}
\usepackage{authblk}
\include{mathpmtc}

\usepackage[pagebackref=true,breaklinks=true,letterpaper=true,colorlinks,bookmarks=false]{hyperref}

\iccvfinalcopy 

\def\iccvPaperID{1946} 

\ificcvfinal\fi
\begin{document}

\title{Single Shot Text Detector with Regional Attention}

\author[1]{Pan He}
\author[2, 3]{Weilin Huang}
\author[3]{Tong He}
\author[1]{Qile Zhu}
\author[3]{Yu Qiao}
\author[1]{Xiaolin Li}
\affil[1]{National Science Foundation Center for Big Learning, University of Florida}
\affil[2]{Department of Engineering Science, University of Oxford}
\affil[3]{Guangdong Provincial Key Laboratory of Computer Vision and Virtual Reality Technology, \newline
Shenzhen Institutes of Advanced Technology, Chinese Academy of Sciences}

\maketitle
\thispagestyle{empty}

\begin{abstract}
We present a novel single-shot text detector that directly outputs word-level bounding boxes in a natural image. We propose an attention mechanism which roughly identifies text regions via an automatically learned attentional map.  This substantially suppresses background interference in the convolutional features, which is the key to producing \textit{accurate} inference of words, particularly at extremely small sizes. This results in a single model that essentially works in a coarse-to-fine manner. It departs from recent FCN-based text detectors which cascade multiple FCN models to achieve an accurate prediction. Furthermore, we develop a hierarchical inception module which efficiently aggregates multi-scale inception features. This enhances local details, and also encodes strong context information,  allowing the detector to work reliably on multi-scale and multi-orientation text with single-scale images.
Our text detector achieves an F-measure of 77\% on the ICDAR 2015 benchmark, advancing the state-of-the-art results in~\cite{Liu2017, Tian2016}. Demo is available at: \url{http://sstd.whuang.org/}.
\end{abstract}
\section{Introduction}

Reading text in the wild has attracted increasing attention in computer vision community, as shown in recent work~\cite{Tian2016, Gupta2016, Zhang2016, Zhu2016, Cho2016, Pan2016_reading}. It has numerous potential applications in image retrieval, industrial automation, robot navigation and scene understanding. Recent work focuses on text detection in natural images, which remains a challenging problem~\cite{Tian2016, Gupta2016, Zhang2016, Zhu2016, Cho2016}. The main difficulty lies in a vast diversity in text scale, orientation, illumination, and font, which often come with a highly complicated background.

\begin{figure}[t!]
   \centering
   \includegraphics[width=8cm, height=6cm]{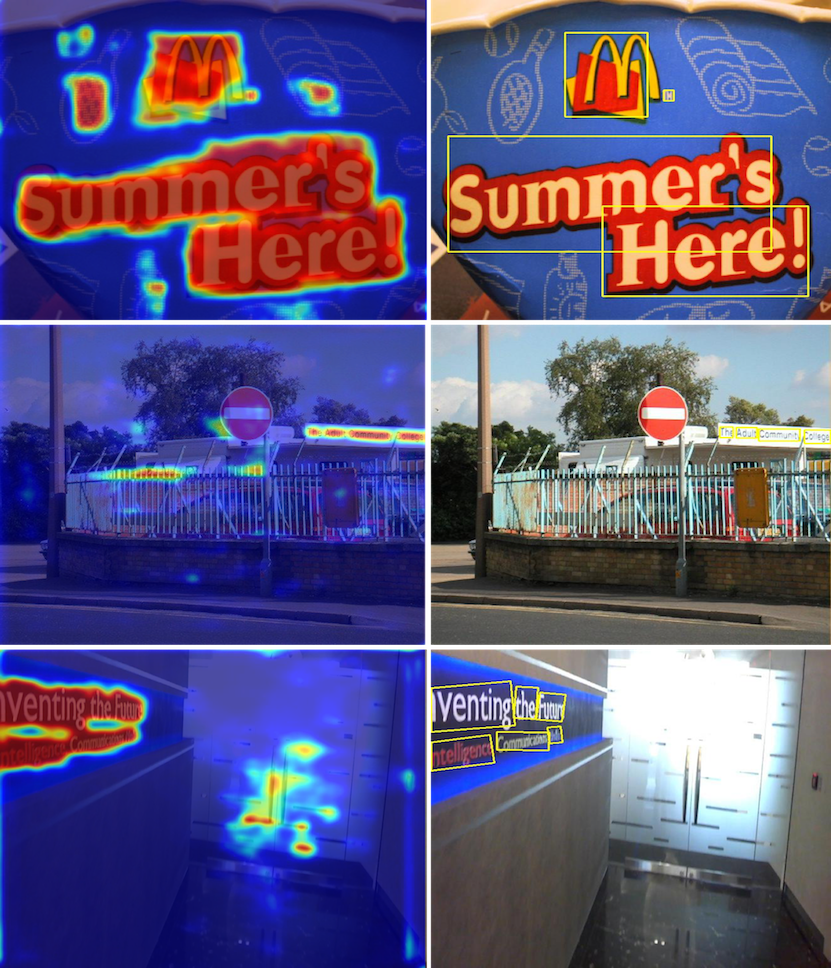}
   \caption{Illustrations of text attention mechanism in the proposed one-shot text detector. Our model automatically learns a rough text regional attention which is used to highlight text information in the convolutional features. This makes it possible to achieve \textit{accurate} word-level predictions in one shot. Text can be precisely separated and accurately predicted at the \textit{word level} in challenging cases.}
   \label{fig:main}
\end{figure}

Previous works in text detection have been dominated by bottom-up approaches~\cite{Epshtein2010,Huang2013, Huang2014, Yin2014,He2016}, which often contain multiple sequential steps, including character or text component detection, followed by character classification or filtering, text line construction and word splitting. Character detection and filtering steps play a key role in such bottom-up approaches. Previous methods typically identify  character or text component candidates using connected component based approaches (e.g., stroke width~\cite{Epshtein2010,Huang2013} or extremal region~\cite{Neumann2012, Huang2014,Yin2014}),  or sliding window methods~\cite{Jaderberg2014, Wang2012,Pan2016_reading}. However, both groups of methods commonly suffer from two main limitations which significantly reduce their efficiencies and performance. First, text detection is built on identification of individual characters or components, making it difficult to explore regional context information. This often results in a low recall where ambiguous characters are easily discarded. It also leads to a reduction in precision, by generating a large number of false detections.  Second, multiple sequential steps make the system highly complicated, and errors are easily accumulated in the later steps.

Deep learning technologies have advanced the performance of text detection considerably~\cite{Tian2016, Gupta2016, Zhang2016, Zhu2016, He2016b, Pan2016_reading}. A number of recent approaches~\cite{Zhang2016,Gupta2016, He2016b, Yao2016} were built on Fully Convolutional Networks (FCN)~\cite{Long2015}, by producing pixel-wise prediction of text or non-text. We refer this group of methods as pixel-based text detectors, which cast previous character-based detections into the problem of text semantic segmentation. This allows them to explore rich regional context information, resulting in a stronger capability for detecting ambiguous text, and also reducing the number of false detections substantially.~\textit{In spite of effectively identifying rough text regions, these FCN-based approaches fail to produce accurate word-level predictions with a single model. The main challenge is to precisely identify individual words from a detected rough region of text.} As indicated in~\cite{Tian2016}, the task of text detection often requires a higher localization accuracy than general object detection. To improve the accuracy, a coast-to-fine detection pipeline was developed, by cascading two FCN models~\cite{He2016b, Zhang2016}. The second FCN produces word- or character-level predictions on a cropped text region detected by the first one. This inevitably increases system complexity: (i) it is highly heuristic yet complicated to correctly crop out regions of texts, words or characters from a  predicted heatmap~\cite{He2016b, Zhang2016}; (ii) multiple bottom-up steps are still required for constructing text lines/words~\cite{Zhang2016, Yao2016}.

Recently, another group of text detectors was developed for direct prediction of text bounding boxes, by extending from the state-of-the-art object detectors, such as Faster R-CNN~\cite{Ren2015} and SSD~\cite{liu2016SSD}. They all aim to predict text boxes directly by sliding a window through the convolutional features \cite{Liu2017, Tian2016, Liao2017, Gupta2016,Zhou2017, Shi2017}, and we refer them as box-based text detectors. \textit{The box-based text detectors are often trained by simply using bounding-box annotations, which may be too coarse (high-level) to provide a direct and detailed supervision, compared to the pixel-based approaches where a text mask is provided.}  This makes the models difficult to learn sufficient word information in details, leading to accuracy loss in one-shot prediction of words, particularly for those small-scale ones. Therefore, they may again come up with multiple post-processing steps to improve the performance.

\subsection{Contributions}
These related approaches inspire current work that aims to directly estimate word-level bounding boxes in one shot. We cast the cascaded FCNs detectors into a single model by introducing a new attention module, which enables a direct mask supervision that explicitly encodes detailed text information in training, and functions on an implicit text region detection in testing. This elegantly bridges the gap between the pixel-based approaches and the box-based text detectors,  resulting in a single-shot model that essentially works in a coarse-to-fine manner. We develop a hierarchical inception module which further enhances the convolutional features. The main contributions of the paper are three-fold.

First, we propose a novel text attention module by introducing a new auxiliary loss, built upon the aggregated inception convolutional features. It explicitly  encodes strong text-specific information using a pixel-wise text mask, allowing the model to learn rough top-down
spatial attention on text regions. This text regional attention significantly suppresses background interference in the convolutional features, which turns out to reduce false detections and also highlight challenging text patterns.

Second, we develop a hierarchical inception module which efficiently aggregates multi-scale inception features. An inception architecture with dilated convolutions \cite{Yu2016}  is applied to each convolutional layer, enabling the model to capture multi-scale image content. The multi-layer aggregations further enhance local detailed information and encode rich context information, resulting in stronger deep features for word prediction.

Third, the proposed text-specific modules are seamlessly incorporated into the SSD framework, which elegantly customizes it towards fast, accurate and single-short text detection.  This results in a powerful text detector that works reliably on multi-scale and multi-orientation text with single-scale inputs. It obtains state-of-the-art performance on the standard ICDAR 2013 and ICDAR 2015 benchmarks, with about $0.13s$ running time on an image of $704 \times 704$.

\begin{figure*}
   \centering
   \includegraphics[width=0.95\textwidth, height=6.5cm]{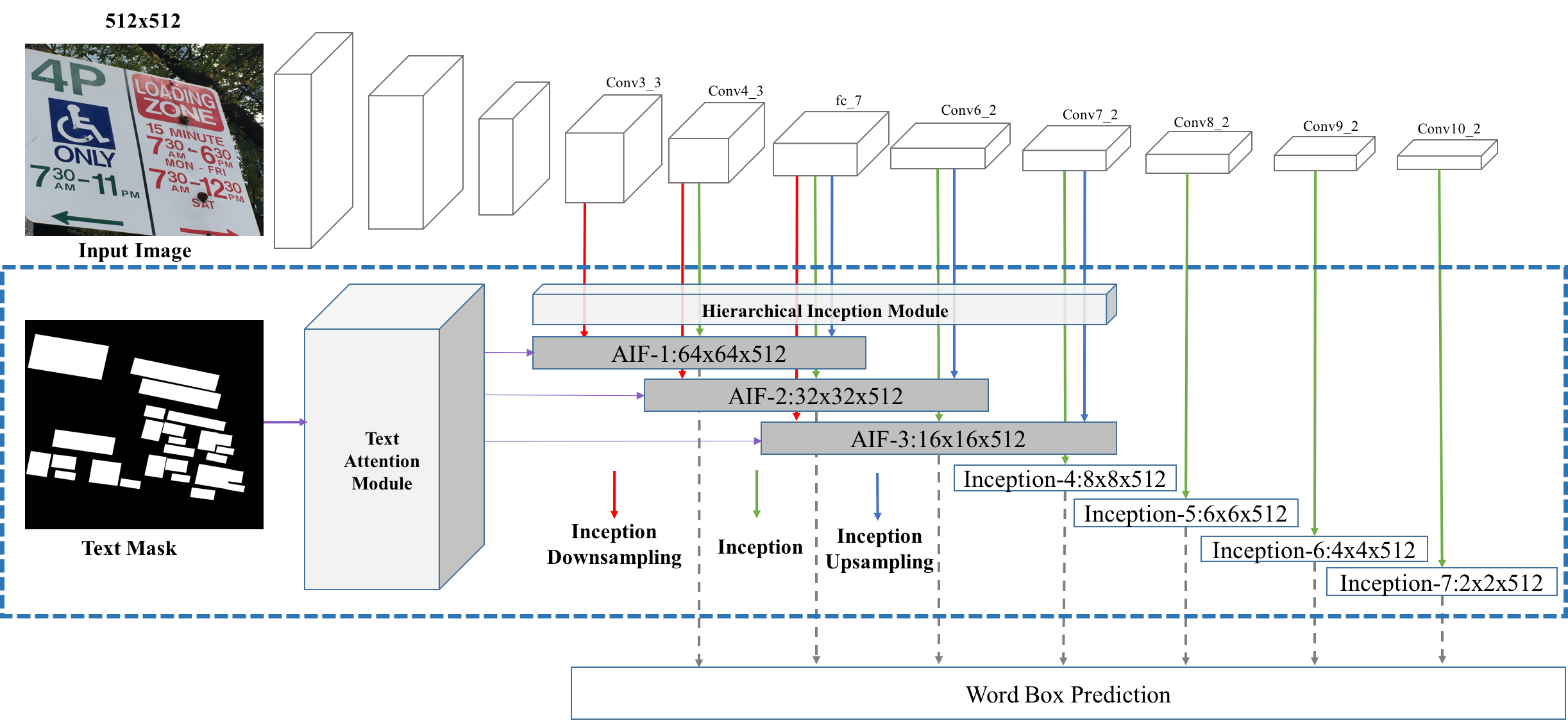}
   \caption{Our single-shot text detector contains three main parts: a convolutional part, a text-specific part, and a word box prediction part. We propose the text-specific part which comprises a Text Attention Module (TAM) and a Hierarchical Inception Module (HIM). The TAM introduces a new pixel-wise supervision of text, allowing the model to automatically learn text attention map which identifies rough text regions. The HIM aggregates multi-layer inception modules, and enhances the convolutional features towards text task.}
   \label{fig:architecture}
\end{figure*}

\section{Related Work}
Previous works on scene text detection mainly focus on bottom-up approaches, which detect characters or text components from images by using hand-crafted features~\cite{Tian2015, Yin2014, Huang2014, Yao2012} or sliding window methods~\cite{Jaderberg2014, Wang2012}. They often involve pixel-level binary classification of text/non-text and generate a text salience map.  Then multiple bottom-up steps are designed to group text-related pixels into characters, which are further formed character pairs, and then text lines. Each step may be followed by a text/non-text filter or classifier. Most of these steps are built on the heuristic or hand-crafted features, e.g., gradient, stroke width ~\cite{Epshtein2010, Huang2013}, covariance descriptor~\cite{Huang2013}, etc. These bottom-up approaches are complicated, and identification of individual characters using low-level features is neither robust nor reliable.

Deep learning technologies have significantly advanced the performance of text detection in the past years~\cite{Tian2016, Gupta2016, Zhang2016, Liao2017, He2016b}. These approaches essentially work in a sliding window fashion, with two key developments: (i) they leverage deep features, jointly learned with a classifier, to enable strong representation of text; (ii) sharing convolutional mechanism~\cite{Long2015,Pan2016_reading} was applied for reducing the computational cost remarkably. With these two improvements, a number of Fully Convolutional Network (FCN)~\cite{Long2015}  based approaches have been proposed~\cite{Zhang2016, He2016b, Yao2016}. They compute pixel-wise semantic estimations of text or non-text, resulting in a fast text detector able to explore rich regional context information. However, these pixel-based text detectors are difficult to provide sufficient localization accuracy by using a single model. Furthermore, accurate segmentation of text from a predicted heatmap is complicated, and often requires a number of heuristic post-processing steps.

Our work also relates to most recent approaches~\cite{Liu2017, Tian2016, Liao2017, Gupta2016, Zhou2017, Shi2017} which are extended from the state-of-the-art object detectors, such as SSD~\cite{liu2016SSD} and Fast R-CNN~\cite{Ren2015}. These approaches all aim to predict text bounding boxes from the convolutional features. Liao \textit{et. al.}~\cite{Liao2017} proposed a TextBox by extending the SSD model for text detection, but their performance is limited by the SSD architecture which is designed for general object detection. Deep Matching Prior Network (DMPNet) was proposed in~\cite{Liu2017}, by introducing quadrilateral sliding windows to handle multi-orientation text. Accurate text localization is achieved by using multi-step coarse-to-fine detection with post adjustments. Tian \textit{et. al.}~\cite{Tian2016} proposed a Connectionist Text Proposal Network (CTPN) which detects a text line by predicting a sequence of fine-scale text components. The CTPN is difficult to work on multi-orientation text and requires bottom-up steps to group text components into text lines. Gupta \textit{et. al.}~\cite{Gupta2016} developed a Fully Convolutional Regression Network (FCRN) which predicts word bounding boxes in an image. However, the FCRN requires three-stage post-processing steps which reduce the efficiency of the system considerably. For example, the post-processing steps take about $1.2s$/image, comparing to $0.07s$/image for bounding box estimations. Our work differs distinctly from these approaches by proposing two text-specific modules. It has a number of appealing properties that advance over these related methods: (i) it is a single-shot detector that directly outputs word bounding boxes, filling the gap between semantic text segmentation and direct regression of word boxes; (ii) it is highly efficient, and does not require any bottom-up or heuristic post-processing step; (iii) it works reliably on multi-orientation text; (iv) it is fast yet accurate, and significantly outperforms those related approaches on the standard ICDAR 2015 benchmark.

\section{Methodology}

We present details of the proposed single-shot text detector, which directly outputs word-level bounding boxes without post-processing, except for a simple NMS. Our detector is composed of three main parts: a convolutional component, a text-specific component, and a box prediction component. The convolutional and box prediction components mainly inherit from SSD detector~\cite{liu2016SSD}. We propose the text-specific component which contains two new modules: a text attention module and a hierarchical inception module. The convolutional architecture of the SSD is extended from the 16-layer VGGNet~\cite{Simonyan2015}, by replacing the fully-connected (FC) layers with several convolutional layers~\cite{liu2016SSD}.
The proposed modules can be easily incorporated into the convolutional component and box prediction component of the SSD, resulting in an end-to-end trainable model, as shown in Fig.~\ref{fig:architecture}. The two text-specific modules elegantly customize the SSD framework towards accurate word detection.  Compared to most recent methods~\cite{Liu2017, Tian2016, Liao2017}, we show experimentally that our particular designs provide a more principled solution that generalizes better.

\subsection{Text Attention Mechanism}
\begin{figure}[t!]
   \centering
  \includegraphics[width=0.45\textwidth, height=5cm]{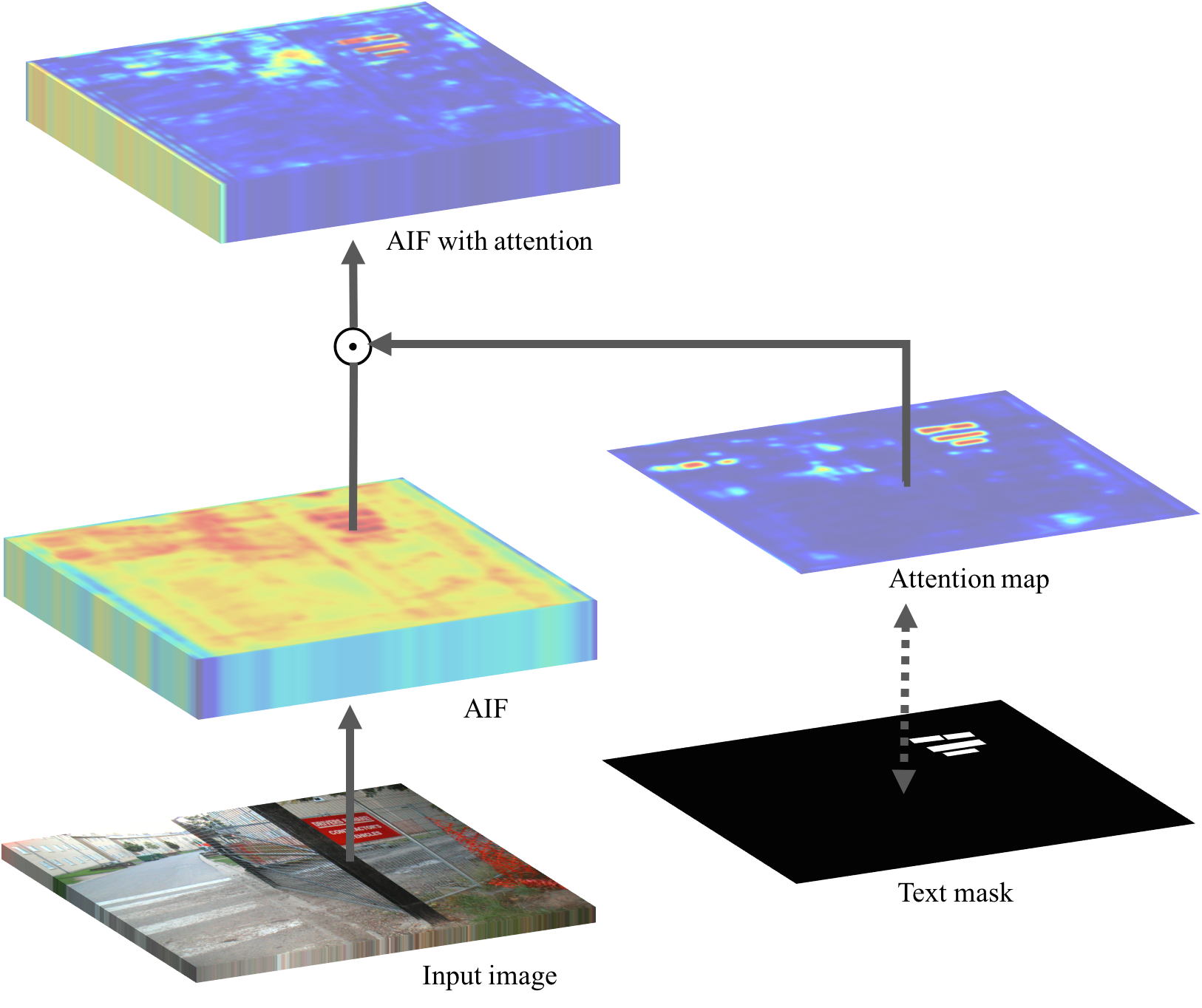}
   \caption{Text attention module. It computes a text attention map from Aggregated Inception Features (AIFs). The attention map indicates rough text regions and is further encoded into the AIFs. The attention module is trained by using a pixel-wise  binary mask of text.}
   \label{fig:attention}
\end{figure}

Our attention module is able to automatically learn rough spatial regions of text from the convolutional features. This attention to text is then directly encoded back into the convolutional features, where text-related features are strongly enhanced by reducing background interference in the convolutional maps, as shown in Fig.~\ref{fig:attention}.

The attention module is built on the Aggregated Inception Feature (AIF) (described in Sect. 3.2). It generates a pixel-wise probability heatmap which indicates the text probability at each pixel location. This probability heatmap is referred as the attention map which has an identical size of an input image and will be downsampled for each prediction layer. The attention module includes two $3\times3$ convolutional layers with pad 1, one deconvolution (upsampling with bilinear operation) layer which connects the AIF to the attention map. Then the attention map is generated by using a softmax activation function on the de-convoluted features. Specifically, given an input image of $512 \times 512$, we get the first-layer AIF features, $F_{AIF_1} \in R^{64\times 64\times 512}$. The attention map, $\alpha^{+} \in R^{512 \times 512}$, is computed as,
\begin{align}
D_{AIF_1} = deconv_{3\times3}(F_{AIF_1}), \\
\bar{D}_{AIF_1} = conv_{1\times1}(D_{AIF_1}), \\
\alpha = softmax(\bar{D}_{AIF_1}).
\end{align}
where $D_{AIF_1} \in R^{512 \times 512 \times 512}$ is the de-convoluted feature maps, which are further projected to 2-channel maps, $\bar{D}_{AIF_1} \in R^{512 \times 512 \times 2}$ using $1\times1$ kernels, followed by a softmax function. Then the positive part of the softmax maps, $\alpha^{+}$, is the attention map, indicating pixel-wise possibility of text. The attention map is further encoded into the AIF by simply resizing it as with spatial size,
\begin{align}
\hat{\alpha}^{+} = resize(\alpha^{+}), \\
\hat{F}_{AIF_1} = \hat{\alpha}^{+} \odot F_{AIF_1}.
\end{align}
where $\hat{\alpha}^{+} \in R^{64 \times 64}$  is the resized attention map, and $\odot$ indicates element-wise dot production across all channel of the AIF maps. $\hat{F}_{AIF_1}$ is the resulted feature maps with encoded text regional attention. The AIFs with and without text attention are shown in Fig.~\ref{fig:attention}, where text information is clearly presented when the attention is encoded.

The text attentional information is learned automatically in the training process. We introduce an auxiliary loss which provides a direct and detailed supervision of text \textit{via} a binary mask that indicates text or non-text at each pixel location.  A softmax function is used to optimize this attention map toward the provided text mask, explicitly encoding strong text information into the attention module. Notice that the proposed attention module is formulated in a unified framework which is trained end to end by allowing for computing back-propagations through all layers.

\begin{figure*}
   \centering
  \includegraphics[width=0.95\textwidth, height=2cm]{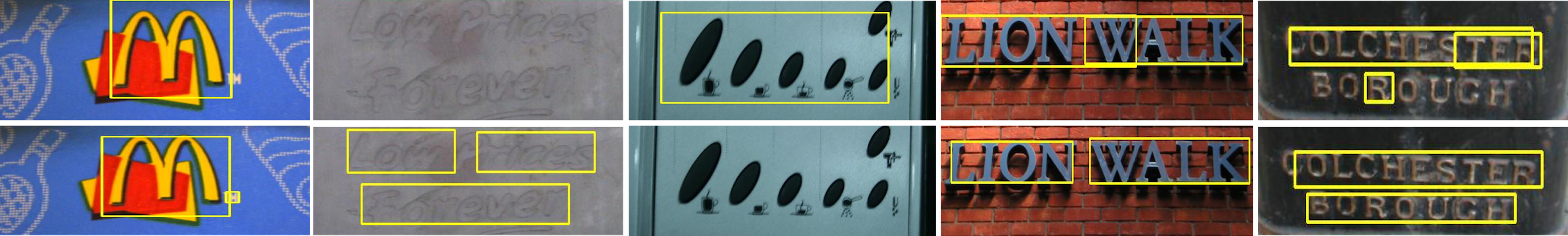}
   \caption{We compare detection results of the baseline model and the model with our text attention module (TAM), which enables the detector with stronger capability for identifying extremely challenging text with a higher \textit{word-level} accuracy. }
   \label{fig:attention_result}
\end{figure*}

The proposed attention module sets this work apart from both previous pixel-wise and box-wise text detectors. It elegantly handles the main limitations of both groups of methods, resulting in an efficient single-shot text detector that produces accurate word-level text detection. Efficiency of the proposed attention module is demonstrated in Fig.~\ref{fig:attention_result}, where detection results by a baseline model and a with-attention model are presented. Obviously, the proposed attention module improves the performance at three aspects: (i) it reduces the number of false detections; (ii) it allows the model to detect more ambiguous texts; (iii) it improves the word-level detection accuracy.

\subsection{Hierarchical Inception Module}
In a CNN model, convolutional features in a lower layer often focus on local image details, while the features in a deeper layer generally capture more high-level abstracted information. The SSD detector predicts object bounding boxes at multi-layer convolutional maps, allowing it to localize multi-scale objects from a single-scale input. Texts often have large variations in scale with significantly different aspect ratios, making the single-scale text detection highly challenging. In recent work \cite{Tian2016}, a layer-wise RNN is incorporated into a convolutional layer,  allowing the detector to explore rich context information thought the whole line. This RNN-based text detector is powerful to detect near-horizontal text lines but is difficult to work reliably on multi-orientation texts.

\begin{figure}
   \centering
   \includegraphics[width=0.45\textwidth, height=3.5cm]{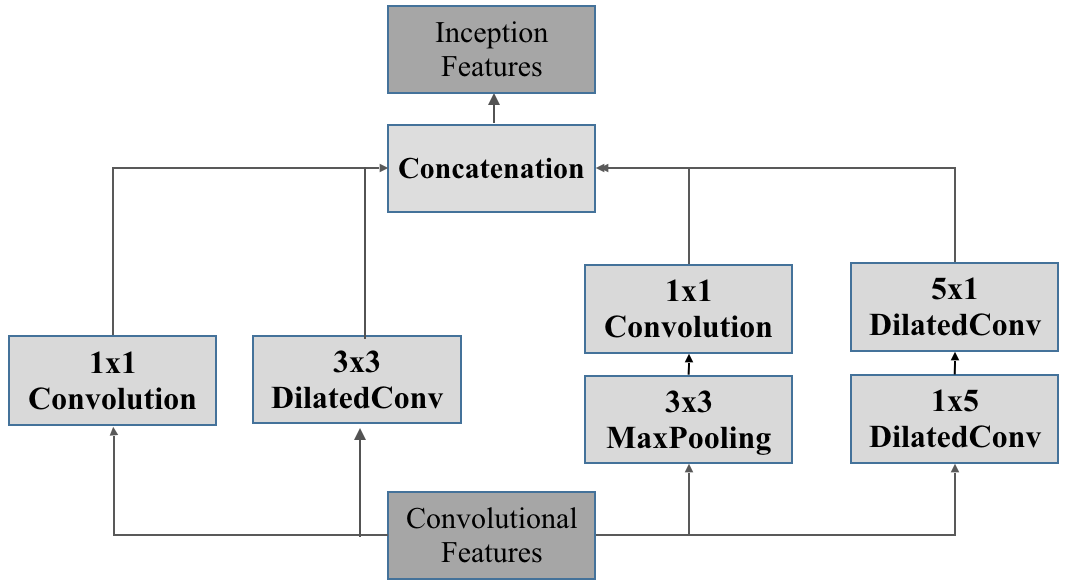}
   \caption{ Inception module. The convolutional maps are processed through four different convolutional operations, with Dilated convolutions \cite{Yu2016} applied.}
   \label{fig:inception}
\end{figure}

Inspired by the inception architecture of GoogleNet \cite{szegedy2015going}, we propose a hierarchical inception module able to aggregate stronger convolutional features. First, we develop a similar inception module which is applied to each convolutional layer, as shown in Fig. \ref{fig:inception}. This allows it to capture richer context information by using multi-scale receptive fields. Second, we aggregate the inception features from multiple layers and generate final Aggregated Inception Features (AIF).

\begin{figure}[!t]
   \centering
  \includegraphics[width=0.45\textwidth, height=5cm]{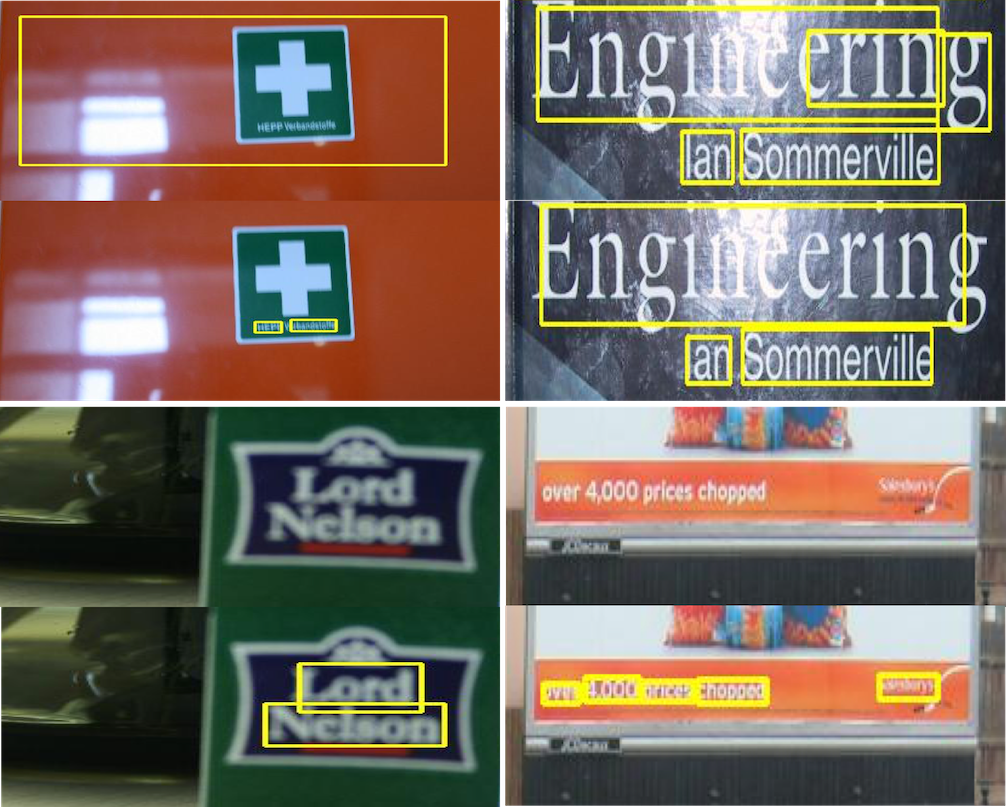}
   \caption{Comparisons of baseline model and Hierarchical Inception Module (HIM) model. The HIM allows the detector to handle extremely challenging text, and also improves \textit{word-level} detection accuracy.}
   \label{fig:hyper}
\end{figure}

Details of inception building block are described in Fig. \ref{fig:inception}.  It is applied to  those convolutional layers which are used to predict word bounding boxes. The convolutional maps in a layer are processed through four different convolutional operations: \textit{$1\times1$-conv}, \textit{$3\times3$-conv}, \textit{$3\times3$-pool} with \textit{$1\times1$-conv}, and \textit{$5\times5$-conv}. The \textit{$5\times5$-conv} is decomposed into \textit{$1\times5$} and \textit{$5\times1$} convolution layers.  Dilated convolutions \cite{Yu2016}, which support an exponential expansion of receptive field without loss of resolution or coverage, are applied. Each convolutional operation reduces the number of feature channels to 128. The final inception features are generated by simply concatenating four 128-channel features, resulting in 512-channel inception features. 
 By using multiple convolutional operations with channel concatenation, the inception features have multi-scale receptive fields and thus can focus on image content in a wide range of scales.

Motivated from HyperNet~\cite{hariharan2015hypercolumns}, we further enhance the convolutional features by aggregating multi-layer inception features, which generates final AIFs at three key convolutional layers, as shown in Fig.~\ref{fig:architecture}. Each AIF is computed by fusing the inception features of current layer with two directly adjacent layers. Down-sampling and up-sampling are applied to the lower layer and higher layer, respectively. These sampling operations ensure same feature resolutions for three inception features, which are combined together using channel concatenation.

The proposed hierarchical inception module is related to the skip architecture developed in \cite{Long2015}, which combines multi-layer convolutional features for handling multi-scale objects. We propose a two-step aggregation approach by leveraging efficient layer-wise inception module. This results in a stronger AIF, with richer local details encoded and a more powerful multi-scale capability. Both improvements are important for text task, allowing our model to identify very small-scale text and work reliably on the multi-scale text which often has a wider range of scales than the general objects.  Compared to layer-wise RNN method developed in \cite{Tian2016}, the AIF encodes more local detailed information and generalizes better to multi-orientation text. The effectiveness of the proposed hierarchical inception module is demonstrated in Fig. \ref{fig:hyper}.

\subsection{Word Prediction Module}
The proposed text-specific modules are directly incorporated into the SSD framework. It can be trained by simply following the SSD, with slight modifications on the box prediction part. As in~\cite{liu2016SSD}, we use a softmax function for binary classification of text or non-text,  and apply the smooth-$l_1$ loss for regressing 5 parameters for each word bounding box, including a parameter for box orientation. Our model predicts $N$ word bounding boxes at each spatial location of the inception or AIF maps. The predictions are computed through all inception and AIF maps shown in Fig. \ref{fig:architecture}.  $N$ is the number of the pre-defined \textit{default} word boxes, which can be pre-computed. A \textit{default} box is defined by its scale and aspect ratio. Since different inception maps have various sizes of receptive fields, the scale of a default box is varied over different inception or AIF maps. Its length is measured by the number of pixels in the input image. To better match the default boxes with ground truth ones, we use three different scales for each layer, and the scales for all inception and AIF layers are listed in Table  \ref{tab:scales}.

\begin{table}
\centering
\caption{Scales and aspect rations of the default box applied for each AIF or Inception layer.}
\label{tab:scales}
\begin{tabular}{l|l|l|l|l|l|l}
\hline
\multicolumn{7}{c}{Scales of Default Box} \\
\hline
\hline   AIF-1       & AIF-2       & AIF-3       & Inc-4       & Inc-5       & Inc-6        & Inc-7     \\ \hline

7.7 & 38.4 & 69.1 & 102.4 & 194.6 & 286.7& 378.9 \\
17.9 & 48.6 & 79.4 & 133.1 & 225.3 & 317.4& 409.6 \\
28.2 & 58.9 & 89.6 & 163.8 & 256.0 & 348.2 & 440.3 \\  \hline\hline
\multicolumn{7}{l}{Aspect Ratios: \{0.5, 1, 2, 3, 5, 7, 9, 11\} }\\\hline
\end{tabular}
\end{table}

Aspect ratios of a default box are set in a wide range (in Table~\ref{tab:scales}), due to the significant variances of them for text. Same aspect ratios are used in all layers. In order to handle multi-orientation text, a scale can be the width or height of a default box. This results in 45 default boxes in total for each layer, which allow the detector to handle large shape variances of text. The center of a default box is identical to the current spatial location of the inception/convolution maps.

\section{Experimental Results}
Our methods are evaluated on three standard benchmarks, the ICDAR 2013~\cite{Karatzas2013}, ICDAR 2015~\cite{Karatzas2015} and COCO-Text dataset~\cite{veit2016cocotext}.
The effectiveness of each proposed component is investigated by producing exploration studies.  Full results are compared with the state-of-the-art performance on the three benchmarks.

\subsection{Datasets and Implementation Details}
\textbf{Datasets.} The\textit{ ICDAR 2013}~\cite{Karatzas2013} consists of 229 training images and 233 testing
images, with word-level annotations provided. It is the standard benchmark for evaluating near-horizontal text detection. We use two standard evaluation protocols: the new ICDAR13 standard \cite{Karatzas2015} and the DetEval~\cite{Karatzas2013}. Our results were obtained by uploading the predicted bounding boxes to the official evaluation system. The \textit{ICDAR 2015} (Incidental Scene Text Challenge 4)~\cite{Karatzas2015} was collected by using Google Glass and it has 1,500 images in total: 1,000 images for training and the remained 500 images for testing. This new benchmark was designed for evaluating multi-orientation text detection. Word-level annotations are provided and each word is labeled by the coordinates of its four corners in a clockwise manner. This dataset is more challenging and has images with arbitrary orientations, motion blur, and low-resolution text. We evaluate our results based on the online evaluation system~\cite{Karatzas2015}. The \textit{COCO-Text}~\cite{veit2016cocotext} is the largest text detection dataset, which has 63,686 annotated images in total: 43,686 for the training and the rest 20,000 for testing.

\textbf{Training datasets.} Our training samples were collected from the training sets of the ICDAR 2013 and ICDAR 2015. We also added images harvested from Internet as the training data and manually labelled them with word level annotation. We have 13, 090 training images in total. We did not use the training data from the COCO-Text.

\textbf{Implementation details.} Our detection network is trained end-to-end by using the mini-batch stochastic gradient descent algorithm, where the batch size is set to 32, with a momentum of 0.9. We initialize our model using the pre-trained model in~\cite{liu2016SSD}. The learning rate is set to 0.001 and is decayed to its $\frac{1}{10}$ after 15, 000 iterations. The model is
trained by fixing the first four convolutional layers, and the training is stopped when the loss no longer decreases.

Data augmentation is used by following the SSD~\cite{liu2016SSD}. We take a similar strategy by randomly sampling a patch from an image and set the minimum Jaccard overlap with ground truth word bounding boxes to $\{0.1, 0.3, 0.5, 0.7, 0.9\}$.
The sampled patches are then resized to $704 \times 704$ and are randomly mirrored with color distortion. 
The NMS threshold and confidence threshold are set to 0.3 and 0.7, respectively. Our methods were implemented using Caffe~\cite{Jia2014}, with  TITAN X GPUs.

\begin{figure*}[t!]
\centering     \includegraphics[width=0.95\textwidth, height=6.7cm]{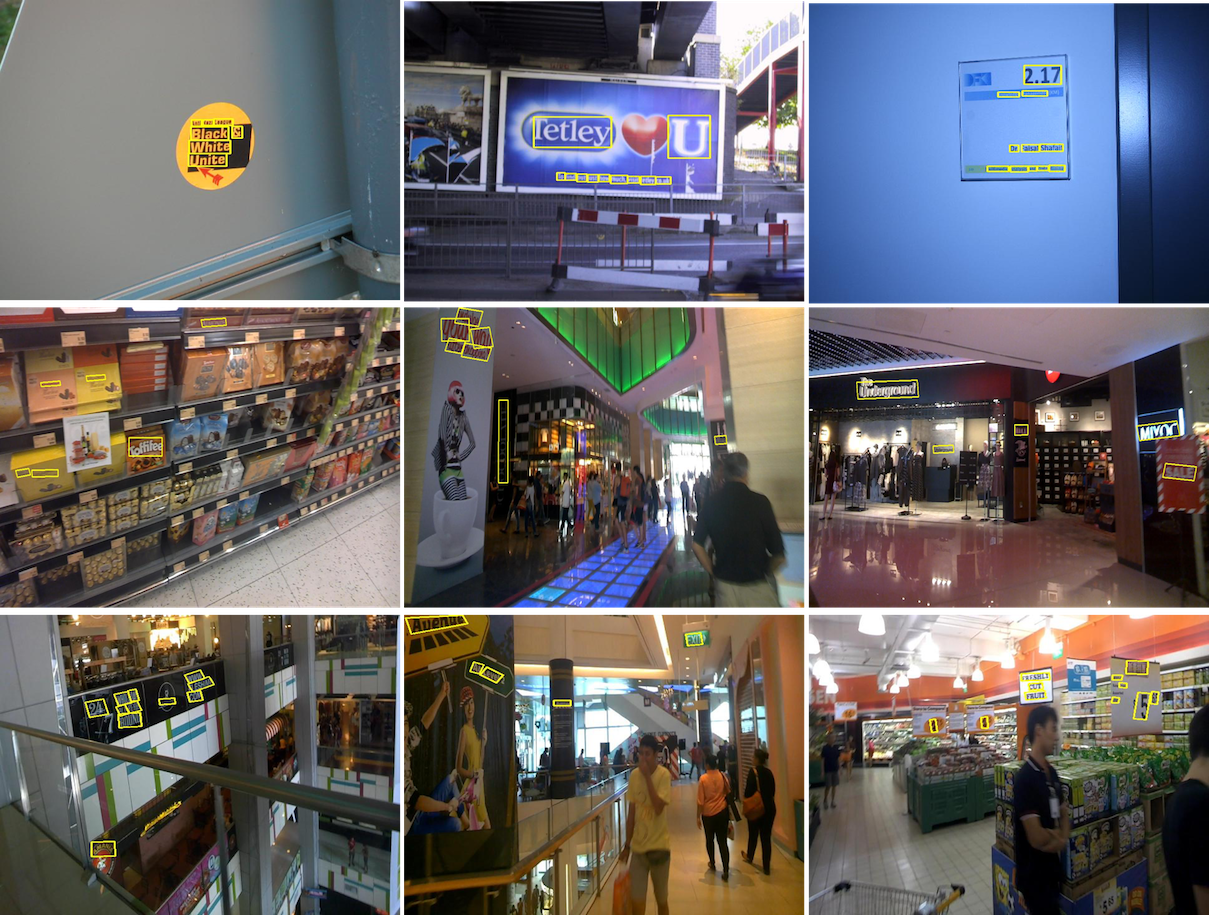}
   \caption{Detection results by the proposed single-shot text detector. }
   \label{fig:qualitative}
\end{figure*}

\subsection{Exploration Study}
We evaluate the effectiveness of the proposed text attentional module (TAM) and hierarchical inception module (HIM) individually. We compare them with our baseline model, which is extended from the SSD framework by simply modifying the word prediction module (as described in Sect. 3.3).  Experimental results on the ICDAR 2013 are compared in Table \ref{tab:exploration}, in the terms of precision, recall and f-measure.
\begin{table}
\centering
\caption{Exploration study on the ICDAR 2013 dataset, based on the new  ICDAR13 standard.}
\begin{tabular}{|l|c|c|c|}

\hline
Method              &        Recall & Precision & F-measure \\ \hline \hline
SSD~\cite{liu2016SSD}                        & 0.60        & 0.80     & 0.68        \\ \hline
Textboxes~\cite{Liao2017}              & 0.74        & 0.86     & 0.80        \\ \hline \hline
Baseline Model        & 0.78        & 0.87     & 0.82        \\ \hline
TAM & 0.83        & 0.87     & 0.85        \\ \hline
HIM & 0.82        & 0.87     & 0.84        \\ \hline

TAM+HIM &           0.86&     0.88&    0.87 \\ \hline
\end{tabular}
\label{tab:exploration}
\end{table}
First, we compare our baseline model with original SSD trained on text data, without any modification of the model. Our baseline model obtains significant performance improvements and outperforms recent TextBoxes \cite{Liao2017}, which is also extended from the SSD. Second, the proposed TAM and HIM both achieve large improvements in recall, indicating that the proposed text-specific modules can increase model accuracy and improve its capability for identifying challenging words, as shown in Fig. \ref{fig:attention_result} and \ref{fig:hyper}.  Third, by incorporating both TAM and HIM into a single model, the final single-shot detector further improves the performance in both recall and precision, obtaining a final F-measure of 0.87. This suggests that both TAM and HIM can compensate for each other in our model. Notice that the performance on the ICDAR2013 is saturated, and the improvement obtained by each component is significant.

\begin{table*}[!t]
\centering
\caption{Comparisons of the state-of-the-art results on the ICDAR 2013 and ICDAR 2015. The results are reported in the terms of Recall (R), Precision (P) and F-measure (F)}
\begin{tabular}{|l|c|c|c|c|c|c|c||l|c|c|c|}
\hline
\multicolumn{8}{|c|}{ICDAR 2013 dataset} & \multicolumn{4}{|c|}{ICDAR 2015 dataset}
\\
\hline
\multirow{2}{*}{Method} & \multicolumn{3}{c|}{ICDAR Standard} & \multicolumn{3}{c|}{DetEval} & \multirow{2}{*}{Time(s)} & \multirow{2}{*}{Method} & \multirow{2}{*}{R} & \multirow{2}{*}{P} & \multirow{2}{*}{F} \\ \cline{2-7}
                        & R  & P  & F & R & P & F &  & & & &                        \\ \hline
SSD~\cite{liu2016SSD} & 0.60 & 0.80  & 0.68 &  0.60 & 0.80  & 0.69 & 0.10 & Deep2Text-MO & 0.32 & 0.50 & 0.39\\
Yin~\cite{Yin2014}  & 0.66 &0.88  & 0.76 & 0.69& 0.89&  0.78&  0.43 &  HUST\_MCLAB & 0.44 & 0.38 & 0.41\\
Neumann~\cite{Neumann2015b}  & 0.72& 0.82 & 0.77 & -  & - & -& 0.40 & AJOU & 0.47 & 0.47  & 0.47\\
Neumann~\cite{Neumann2015}  & 0.71 & 0.82 & 0.76 & -  & - & -& 0.40 & NJU-Text & 0.36 & 0.73  & 0.48\\
FASText~\cite{Busta2015}  & 0.69 & 0.84 & 0.77& -  & - & -& 0.15 & CASIA USTB & 0.40 & 0.62  & 0.48  \\
Zhang~\cite{Zhang2015}  & 0.74 & 0.88  & 0.80 &0.76  & 0.88 & 0.82 & 60 & StradVision1 & 0.46 & 0.53  & 0.50 \\
TextFlow~\cite{Tian2015}  & 0.76 & 0.85  & 0.80 &- &-&- & 0.94 & StradVision2 &  0.37 & 0.77  & 0.50 \\
Text-CNN~\cite{He2016}  & 0.73 &0.93  & 0.82 & 0.76& \textbf{0.93} & 0.84 & 4.6 & MCLAB\_FCN~\cite{Zhang2016} & 0.43 & 0.71  & 0.54   \\
FCRN~\cite{Gupta2016} & 0.76 & \textbf{0.94}  & 0.84 & 0.76 & 0.92  & 0.83 & 1.27 & CTPN~\cite{Tian2016} & 0.52 & 0.74 & 0.61 \\
TextBoxes~\cite{Liao2017} & 0.74 & 0.86 & 0.80 & 0.74 & 0.88 & 0.81 & 0.09 & Yao \textit{et. al.}\cite{Yao2016} & 0.57 & 0.72 & 0.64 \\
CTPN~\cite{Tian2016} & 0.73 & 0.93 & 0.82 & 0.83 & \textbf{0.93} & \textbf{0.88}& 0.14 & DMPNet~\cite{Liu2017} & 0.68 & 0.73 & 0.71\\
\hline
\hline
Proposed method  & \textbf{0.86} & 0.88 &  \textbf{0.87} & \textbf{0.86} & 0.89 & \textbf{0.88} &  0.13 & Proposed method  & \textbf{0.73} & \textbf{0.80} & \textbf{0.77} \\  \hline
\end{tabular}
\label{tab:icdar}
\end{table*}

\begin{table}[h]
\centering
\caption{Comparisons on the COCO-text dataset}
\begin{tabular}{|l|c|c|c|}
\hline
\multicolumn{4}{|c|}{COCO-text dataset} \\ \hline
Method                      & Recall & Precision & F-score \\ \hline
Baseline model C          & 0.05        & 0.19     & 0.07        \\
Baseline model B           & 0.11        & 0.90     & 0.19        \\
Baseline model A                         & 0.23        & 0.84     & 0.36        \\
Yao~\cite{Yao2016}     & 0.27       & 0.43     & 0.33        \\
Proposed method     &  \textbf{0.31}&     \textbf{0.46} &    \textbf{0.37} \\ \hline
\end{tabular}
\label{tab:coco}
\end{table}

\subsection{Comparisons with state-of-the-art methods}
\textbf{Qualitative results.} Detection results on a number of very challenging images are demonstrated in Fig.~\ref{fig:qualitative}, where our detector is able to correctly identify many extremely challenging words, some of which are even difficult for human. It is important to point out that the word-level detection by our detector is particularly accurate, even for those very small-scale words that are closed to each other.

\textbf{On the ICDAR 2013.} We compare our performance against recently published results in Table \ref{tab:icdar}. All our results are reported on single-scale test images. On the ICDAR 2013 which is designed for near-horizontal text detection, our detector achieves the state-of-the-art performance on both evaluation standards. By using the ICDAR 2013 standard, which focuses on word-level evaluation, our method obtains an F-measure of 0.87, outperforming all other methods compared, including recent FCRN~\cite{Gupta2016}, CTPN~\cite{Tian2016}, and TextBoxes~\cite{Liao2017}. By using the DetEval standard, our method is comparable with the CTPN, and again has substantial improvements over the others. It has to point out that the DetEval standard allows for text-line level evaluation as well,  and is less strict than the ICDAR 2013 standard. Our method encourages a more accurate detection in word level, resulting in better performance in the ICDAR 2013 standard.

In addition, we further evaluate our method for end-to-end word spotting on the ICDAR 2013 by directly combining our detector with recent word recognition model presented in~\cite{Pan2016_reading}. We obtain an accuracy of 0.83 on \textit{generic} case, which is comparable to recent results: 0.79 in~\cite{Jaderberg2015} and 0.85 in~\cite{Gupta2016}.

\textbf{On the ICDAR 2015.} The ability to work on multi-orientation text is verified on the ICDAR 2015. Our method obtains an F-measure of 0.77, which is a remarkable improvement over 0.61 achieved by CTPN \cite{Tian2016}. This clearly indicates that our method generalizes much better to multi-orientation text than the CTPN.  On the ICDAR 2015, our method got the state-of-the-art performance in terms of recall, precision, and F-measure, surpassing recently published results in \cite{Liu2017} (0.71 F-measure) by a large margin.

In all implementations, our method obtains a higher recall than the others. This suggests that the proposed text-specific modules enable our detector with stronger capability for detecting extremely challenging text in a high word-level accuracy, including the very small-scale, significantly-slant or highly-ambiguous words, as shown in Fig. \ref{fig:qualitative}.

\textbf{On the COCO-Text.} We further evaluate our detector on the COCO-Text~\cite{veit2016cocotext}, which is a large-scale text dataset. We achieve state-of-the-art performance with an F-score of 0.37, which improves recent state-of-the-art result~\cite{Yao2016} slightly. This demonstrates strong generalization ability of our method to work practically on large-scale images in the unseen scenarios.  Again, our method achieves a significantly higher recall than all compared approaches.

\textbf{Running time.} We compare running time of various methods on the ICDAR 2013 (in Table \ref{tab:icdar}). Our detector achieves running time of 0.13s/image using a single GPU, which is slightly faster than CTPN using 0.14s/image. It is sightly slower than TextBoxes \cite{Liao2017}, but has substantial performance improvements. In addition, the TextBoxes was not tested on the multi-orientation text. Besides, FCRN \cite{Gupta2016} predicts word boxes at 0.07s/image, but it takes 1.2s/image for post-processing the generated boxes. All recent CNN-based approaches are compared on GPU time, which has become the main stream with success of deep learning technologies.  The fastest CPU based text detector is FASText \cite{Busta2015}, using 0.15s/image. But its performance is significantly lower than recent CNN-based approaches.

\section{Conclusion}
We have presented a fast yet accurate text detector that predicts word-level bounding boxes in one shot. We proposed a novel text attention mechanism which encodes strong supervised information of text in training. This enables the model to automatically learn a text attentional map that implicitly identifies rough text regions in testing, allowing it to work essentially in a coarse-to-fine manner. We developed a hierarchical inception module that efficiently aggregates multi-scale inception features. This further enhances convolutional features by encoding more local details and stronger context information. Both text-specific developments result in a powerful text detector that works reliably on the multi-scale and multi-orientation text. Our method achieved new state-of-the-art results on the ICDAR 2013, ICDAR 2015 and COCO-Text benchmarks.

\section*{Acknowledgement}
This work is supported in part by National Science Foundation (CNS-1624782, OAC-1229576, CCF-1128805), National Institutes of Health (R01-GM110240), Industrial Collaboration Project (Y5Z0371001), National Natural Science Foundation of China (U1613211, 61503367) and Guangdong Research Program (2015B010129013, 2015A030310289).

{\small
\bibliographystyle{ieee}
\bibliography{sstd}
}

\end{document}